\documentclass[journal=jacsat,manuscript=article]{achemso}

\usepackage{chemformula} 
\usepackage[T1]{fontenc} 
\usepackage{hyperref}
\usepackage{multirow}
\usepackage{booktabs}

\author{Junren Li}
\affiliation[Peking University]{Peking University, College of Chemistry and Molecular Engineering, No. 5 Yiheyuan Road, Beijing, China}
\author{Lei Fang}
\affiliation[Microsoft]{Microsoft Corporation, Building 2, No. 5 Dan Ling Street, Beijing, China}
\email{leifa@microsoft.com}
\author{Jian-Guang Lou}
\affiliation[Microsoft]{Microsoft Corporation, Building 2, No. 5 Dan Ling Street, Beijing, China}

\title[Retro-BLEU]
  {Retro-BLEU: Quantifying Chemical Plausibility of Retrosynthesis Routes through Reaction Template Sequence Analysis}

\abbreviations{IR,NMR,UV}

\begin{document}

\begin{tocentry}

\includegraphics{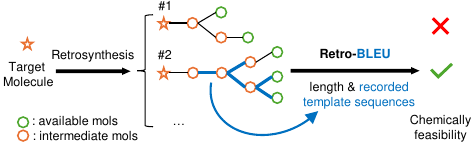}

\end{tocentry}

\begin{abstract}
Computer-assisted methods have emerged as valuable tools for retrosynthesis analysis.
However, quantifying the plausibility of generated retrosynthesis routes remains a challenging task.
We introduce Retro-BLEU, a statistical metric adapted from the well-established BLEU score in machine translation, to evaluate the plausibility of retrosynthesis routes based on reaction template sequences analysis.
We demonstrate the effectiveness of Retro-BLEU by applying it to a diverse set of retrosynthesis routes generated by state-of-the-art algorithms and compare the performance with other evaluation metrics.
The results show that Retro-BLEU is capable of differentiating between plausible and implausible routes.
Furthermore, we provide insights into the strengths and weaknesses of Retro-BLEU, paving the way for future developments and improvements in this field.
\end{abstract}

\section{Introduction}\label{sec1}

Retrosynthesis analysis plays a crucial role in the design and discovery of new chemical compounds\cite{retrosynthesis_1988}. 
A retrosynthesis route usually consists of multiple chemical reactions, decomposing the target molecule into commercially available starting materials in a step-by-step manner\cite{review2022}.
Recently, deep learning-based approaches have substantially expedited the process of retrosynthesis planning, known as Computer-Assisted Synthesis Planning (CASP)\cite{wirsreview, zhong2023review}.
For example, ASKCOS\cite{coley2019askcos}, an open-source platform, can easily generate hundreds of retrosynthesis routes for a given target molecule.
However, it is important to note that not every generated route is guaranteed to be feasible, as the predicted reactants may not yield the expected product in actual lab scenarios\cite{kim2021valid}. 
Therefore, it is crucial to develop metrics to assess the validity and plausibility of these model-generated routes.

Existing metrics to evaluate the retrosynthesis routes can be broadly classified into two primary categories: 
\begin{itemize}
    \item Metrics based on intrinsic properties of generated routes, e.g., route length, reactants price\cite{costeffective2019}, or the coverage of the starting materials in recorded routes\cite{liu2023fusionretro}. While these metrics provide valuable information, they cannot capture the chemical plausibility or practicality of a given route\cite{review2022, costeffective2019}. For example, protection and deprotection steps are essential for obtaining the target product by preventing undesired reactions, which increases the route length.
    \item Metrics based on trained models, e.g., reaction cost\cite{chen2020retro}, which calculates a route-level probability score by multiplying the probabilities of each reaction step. A typical planning system generally consists of a single-step retrosynthesis model\cite{dai2019gln, schwaller2019moltrans, fang2023single} and a multi-step searching algorithm\cite{segler2018MCTS, lin2020MCTS, chen2020retro}. The probabilities generated by single-step models represent the model's confidence derived from the underlying training data\cite{li2023retroranker}, they do not correspond to actual reaction probabilities, which are influenced by various factors such as reaction kinetics and the presence of catalysts. Moreover, the model's performance degrades when the size of the template library is increased, resulting in less reliable probabilities~\cite{datasetdiscussion2019}, and the metric is also affected by the route length, because a route comprising more steps typically exhibits a lower cumulative probability.
\end{itemize}

The goal of retrosynthesis planning is to provide valid routes for synthesis design. 
A valid route indicates that all reactions of the route can be performed in the real-world lab scenario, instead of simply applying reaction templates to arbitrary chemical environments.
Nonetheless, the metrics mentioned above cannot determine the route validity, 
which leaves a gap between current CASP programs and actual laboratory experiments. 
In order to accurately determine if a reaction can take place, a theoretical evaluation or wet-lab experiment is indispensable. 
Such assessments necessitate substantial computational resources (starting from first principles, which are typically challenging to compute precisely) or involve considerable labor costs.
These challenges motivate us to approach the problem from a statistical perspective, seeking statistical measures correlated with chemical plausibility, and thus enabling us to quantify the feasibility of chemical reaction routes.  

\begin{figure*}[htbp]
\begin{center}
\includegraphics[width=0.9\textwidth]{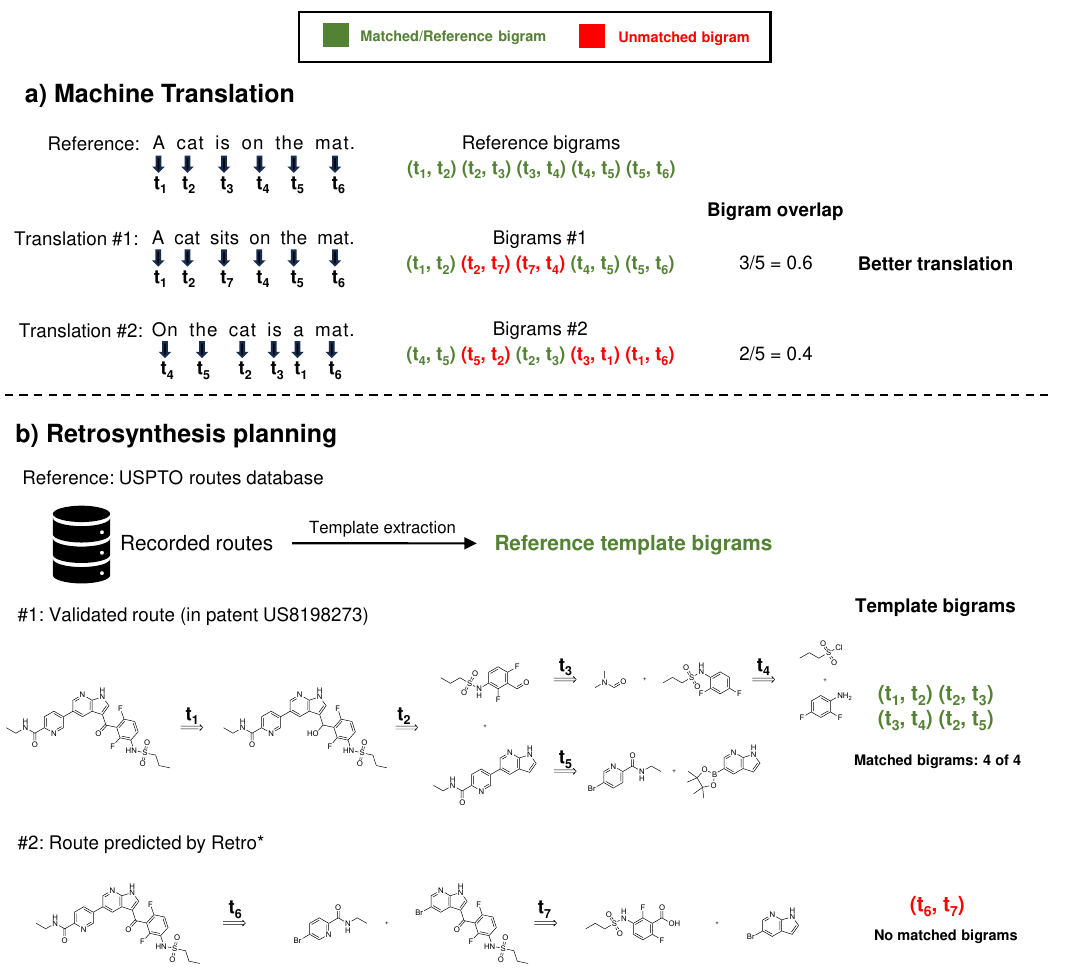}
\caption{An comparative view of evaluation in machine translation and retrosynthesis planning using bigram overlap: 
a) in machine translation, the BLEU-2 score, which can be considered as bigram overlap in this case, can be used to select high-quality translation. 
b) in retrosynthesis planning, template bigrams overlap can be used to select chemically plausible routes. 
}
\label{fig1}
\end{center}
\end{figure*}

In natural language processing (NLP), widely accepted evaluation metrics for tasks such as machine translation or text generation/summarization include Bilingual Evaluation Understudy (BLEU, as we demonstrated in Figure 1)\cite{papineni2002bleu} and Recall-Oriented Understudy for Gisting Evaluation (ROUGE)\cite{lin2004rouge}, and they focus on precision and recall when evaluating with human translation, respectively.
Both BLEU and ROUGE rely on the concept of n-grams to compute the overlap between generated text and the reference text.
n-grams are sequences of ``n'' consecutive words. For example, unigrams represent single words, bigrams represent two consecutive words, and so on.
Drawing an analogy to NLP, retrosynthetic routes (typically represented as trees) can also be considered collections of reaction sequences, as we demonstrated in Figure 1b).
Each sequence corresponds to a specific reaction pathway connecting the target product to leaf nodes, which represent a set of individual starting materials.
Similar to how consecutive words in a sentence often exhibit semantic correlations, consecutive reactions in validated synthesis routes also demonstrate interrelated synthetic strategies, reflecting the underlying logic and coherence in chemical transformations \cite{molga2022iterative, Tactical2020}.
For instance, the nitro group can be easily introduced into an aromatic ring, then reduced to an amine, followed by other substitution reactions, as a simple example of sequential reactions.
Since there is no absolute best route for retrosynthesis planning, the precision of sequential reactions is more important than recall. 
This motivates us to modify the BLEU score for the scenario of retrosynthesis, resulting in Retro-BLEU. 
The key difference between the basic BLEU score and Retro-BLEU lies in the data being analyzed. 
While BLEU deals with text, Retro-BLEU is designed for reaction sequences, which can be obtained from retrosynthetic routes.
In this context, n-grams represent sequences of ``n'' consecutive reactions (or consecutive reaction templates, as we will discuss later) instead of words. 
This adaptation allows us to apply the concept of n-grams from natural language processing to the domain of retrosynthetic routes, enabling a more relevant and meaningful comparison between generated routes and known synthesis routes.
By calculating the precision of matching reaction n-grams between generated routes and known synthesis routes, Retro-BLEU offers a quantifiable approach to assess the quality of generated synthetic pathways.

\section{Dataset and Methods}\label{sec2}

\subsection{n-gram overlap analysis}\label{subsec2}

We first study the n-gram overlap of the reaction sequences in retrosynthesis routes, which will be further used in Retro-BLEU.
In NLP, the overlap is calculated over n-grams in the reference text to measure the semantic correlations between the text generated by a model and the reference text.
While in retrosynthesis, our goal is to measure the feasibility of a synthesis route, the n-gram overlap is determined over n-grams in all known experimentally validated synthesis routes.
The rationale behind this is that if a proposed route shares a significant number of n-grams with known and successful synthesis routes, it is more likely to be chemically plausible and experimentally viable.

We employ two datasets, the PaRoutes~\cite{genheden2022paroutes} dataset and the Retro$^*$-190~\cite{chen2020retro} dataset, to determine if there is a noticeable relationship between the n-gram overlap found in model-generated routes and patent routes.
\begin{itemize}
\item \textbf{PaRoutes}: Following PaRoutes~\cite{genheden2022paroutes}, we collected 457,447 experimentally validated routes from the US Patent and Trademark Office (USPTO) dataset~\cite{lowe2012extraction}. 
PaRoutes also provides two sets of 10,000 diverse, non-overlapping routes with a depth of at most 10 reactions: set-n1 and set-n5.
The difference is the number of routes extracted from each patent before checking for overlapping routes: one route for set-n1 and five routes for set-n5, please refer to PaRoutes\cite{genheden2022paroutes} for details. 
Due to space limitations, we mainly report the results on set-n5 because the results on set-n1 are similar.
We constructed the known n-grams from the patent dataset, excluding those patents containing the corresponding 10,000 target instances.
As a result, the remaining patents are denoted as the corresponding ``known routes''.
This approach was taken to mimic the scenario when evaluating retrosynthesis routes for new targets, ensuring a fair and unbiased comparison.
In addition, we generated 2,958,811 routes for set-n5 molecules using Monte Carlo Tree Search (MCTS)\cite{segler2018MCTS} and 2,799,023 routes for set-n5 molecules using Retro$^*$~\cite{chen2020retro} with AiZynthFinder~\cite{aizynthfinder}, i.e., for each target molecule, we generated approximately 300 routes. 
We use the default parameter settings in AiZynthFinder and employ the top-50 predictions from the single-step model in each step.

\item \textbf{Retro$^*$-190}: Retro$^*$-190~\cite{chen2020retro} is a collection of 190 challenging target molecules specifically designed to test the performance of retrosynthesis search algorithms.
Retro$^*$~\cite{chen2020retro} provided the shortest route for each target by concatenating reactions from various patents until the starting materials are available in \textit{eMolecules}\footnote{https://www.emolecules.com/products/building-blocks}, which are considered as patent routes.
It should be noted that these routes are pseudo-routes because their corresponding reaction sequences may not be chemically logical.
We employ the results of several state-of-the-art search algorithms as model-generated routes to compare the n-gram overlap with known routes.
The 299,902 training routes from Retro$^*$ are considered as known routes to build the known n-grams.
\end{itemize}

On each dataset, we collected n-consecutive reactions (with n ranging from 2 to 4) from the set of corresponding known synthesis routes to construct the known reaction n-gram sequences. 
We utilized the SMILES (Simplified Molecular-Input Line-Entry System) representation for these reactions, as the canonical SMILES of each molecule is unique, allowing for efficient identity checking between tuples.
For example, if a route is 6 steps long, we would take the first four reactions, the middle four reactions, and the last four reactions as three 4-grams.
For each of the tested route, which includes both patent routes and model-generated routes, we extracted n-consecutive reactions and computed their overlap ratio with the known reaction n-grams(routes from the same patents are excluded when constructing dataset), then we averaged the ratio across all routes to obtain the overall overlap ratio.
To be specific, the fraction of n-gram overlap is calculated as follow:
\begin{equation} \label {eq0}
			f_{n}(r) = \frac{N_{\rm recorded}(r)}{N_{\rm total}(r)}
\end{equation}
where $r$ denotes the route, $N_{\rm recorded}$ is the number of recorded n-grams in the route $r$, and $N_{\rm total}$ is the total number of n-grams in the route $r$.

We also calculated the coverage, which is the average ratio of routes having n-grams, e.g., routes shorter than 3 steps do not contain any trigram reaction sequences.

\begin{table*}[htbp]
	\begin{center}
	\caption{n-gram (reaction/templates) overlap ratio for patent routes and model-generated routes.}\label{tab1}
 \resizebox{\linewidth}{!}{
		\begin{tabular}{lcccccccccc}	
\toprule		
n-grams                                                         & \multicolumn{3}{c}{n=2}           & \multicolumn{3}{c}{n=3}           & \multicolumn{3}{c}{n=4}         & \multirow{2}{*}{Avg. length}\\
ratio category                                                  & reaction & template & coverage    & reaction & template & coverage    & reaction & template & coverage  &                             \\
\midrule
PaRoutes set-n5                                         & 49.0\%   & 70.0\%	  & 100\%       & 47.6\%   & 52.3\%   &	92.9\%      & 46.4\%   & 48.7\%   & 51.3\%    & 3.84                        \\
\midrule
MCTS-1\textsuperscript{[a]}                                     &  7.9\%   & 24.0\%	  & 94.3\%      &  7.3\%   &  5.5\%   &	56.7\%      &  7.4\%   &  1.6\%   & 24.8\%    & 3.75                        \\
MCTS-10                                                         &  3.9\%   & 21.2\%	  & 99.0\%      &  2.2\%   &  2.9\%   &	76.7\%      &  1.0\%   &  0.5\%   & 41.2\%    & 4.50                        \\
MCTS-all                                                        &  1.4\%   & 29.1\%	  & 100\%       &  0.3\%   &  3.4\%   &	98.6\%      &  0.1\%   &  0.3\%   & 92.6\%    & 8.65                        \\
Retro$^*$-1                                                     &  4.5\%   & 14.9\%	  & 94.3\%      &  3.4\%   &  2.1\%   &	57.5\%      &  2.4\%   &  0.7\%   & 22.4\%    & 3.23                        \\
Retro$^*$-10                                                    &  3.1\%   & 16.1\%	  & 99.0\%      &  1.7\%   &  1.6\%   &	75.0\%      &  1.0\%   &  0.4\%   & 35.2\%    & 3.79                        \\
Retro$^*$-all                                                   &  2.3\%   & 25.1\%	  & 100\%       &  0.6\%   &  2.4\%   &	98.0\%      &  0.1\%   &  0.2\%   & 84.6\%    & 5.38                        \\
\midrule
\midrule
Retro$^*$-190\cite{chen2020retro}\textsuperscript{[b]}	    & 10.1\%   & 42.9\%	  & 100\%       & 4.6\%    & 28.9\%   &	90.5\%      & 1.5\%    & 21.5\%   & 77.9\%    & 6.67                        \\
\midrule
Retro$^*$(165)\cite{chen2020retro} 	                            &  6.0\%   & 31.2\%	  & 100\%       & 3.4\%    & 16.9\%   &	88.5\%      & 2.1\%    & 14.9\%   & 75.2\%    & 6.35                        \\
Retro$^*$+(183)\cite{retrostarplus}\textsuperscript{[c]} 	    &  3.6\%   & 29.5\%	  & 100\%       & 1.6\%    & 13.6\%   &	90.7\%      & 0.9\%    & 10.2\%   & 79.8\%    & 6.82                        \\
EG-MCTS(183)\cite{egmcts} 	                                    &  1.2\%   & 13.9\%	  & 100\%       & 0.5\%    & 5.2\%    &	90.1\%      & 0.1\%    & 3.1\%    & 72.7\%    & 5.69                        \\
RetroGraph(189)\cite{xie2022retrograph} 	                    &  2.1\%   & 20.1\%	  & 100\%       & 0.9\%    & 7.6\%    &	90.5\%      & 0.3\%    & 4.5\%    & 75.1\%    & 6.40                        \\
\bottomrule	
	\end{tabular}
 }
	\end{center}
\footnotesize{\textsf{
[a] The numbers represent how many routes are used in the evaluation, \textit{i.e.,} top-1 predicted routes, top-10 predicted routes, and all predicted routes (approximately 300 for each target).
[b] The number in the parentheses denotes the solved routes among the 190 targets.
\textsf{[c]} We use the variant of Retro$^*$+ without value functions.}
}
\end{table*}

As shown in Table~\ref{tab1}, on PaRoutes set-n5, nearly half of the reaction n-grams are recorded/known when evaluating the patent routes.
This observation suggests that a significant portion of the reaction sequences in set-n5 patent routes overlap with those found in known synthesis routes, indicating that chemists often rely on familiar and well-understood reaction sequences when designing new synthesis strategies.
However, the overlap ratio declined to less than 10\% on generated routes for both MCTS and Retro$^*$. 
On Retro$^*$-190, which is a quite challenging dataset, the overlap ratio of the pseudo-routes decreases to approximately 10\% at the bigram level, because these routes contain many unobserved reactions from the training data.
This decrease can be attributed to the sparsity of reaction sequences\cite{stocker2020chemspace}.
When considering reactions as individual tokens, the space formed by continuous n-grams is extremely sparse, because encountering unseen reactions is inevitable during the synthesis of novel molecules.
Nonetheless, this does not imply that we should regard unseen reactions as invalid choices. 

The sparsity of the reaction space encourages us to develop a more flexible evaluation of generated routes, emphasizing the underlying chemical transformations. 
In the context of chemical reactions, templates can be considered as an induction and generalization form of reactions.
Therefore, we conducted a similar analysis on template sequences, using the same approach as in analyzing the overlap of reaction sequences. 
The reaction templates are extracted with the rxnutils\cite{kannas2022rxnutils} and we use SMARTS (SMILES Arbitrary Target Specification) strings to demonstrate these templates.
We tested the reaction templates with radii ranging from 0 to 2. 
Herein, we present the results for a radius of 1, results for other radii can be found in Supporting Table 2, indicating that using a radius of 1 is an optimal choice for evaluating template sequences.
As shown in Table~\ref{tab1}, the patent-extracted routes on PaRoutes set-n5 have a significant portion of known consecutive template sequences, much higher than using reaction sequences.
Meanwhile, the overall template sequence overlap ratio is considerably higher than the reaction sequence.
Similarly, the test routes on Retro$^*$-190 have 42.9\% of recorded template bigrams, while model-generated routes exhibit lower overlaps.

It is important to note that coverage is closely related to the average route length. 
When more generated routes are examined for each target, the average length increases, resulting in higher coverage. 
However, only the bigram coverage consistently remains near 100\%. 
Taking the coverage into account, we propose that the bigram overlap ratio should be considered when assessing the chemical plausibility.
Furthermore, it should be noted that the template bigram overlap ratio increases when the average route length increases. 
This might be due to randomly paired sequences as the route extends, which may contain unproductive steps, such as performing unnecessary protection before converting functional groups.
This observation implies that route length should also be considered when evaluating the plausibility of generated routes.

\subsection{Retro-BLEU metric}
Based on n-gram overlap analysis, we propose Retro-BLEU as a method for evaluating the plausibility of retrosynthetic routes from a statistical perspective, although we acknowledge that the ultimate confirmation of a reaction's feasibility relies on wet lab experiments.
The n-gram overlap with known routes between validated routes and generated routes demonstrates a noticeable difference.
Considering that the overlap ratio could be affected by route length, we integrated both the route length and overlap ratio of template n-grams with known routes in Retro-BLEU score: 
\begin{equation} \label {eq1}
			Score_{\rm Retro-BLEU}(r) = \exp{\frac{L}{\max(L, len(r))}} + \exp{f_{n}(r)}
\end{equation}
where $L$ is a hyperparameter, $len(r)$ is the number of reaction steps in route $r$, and $f_{n}(r)$ is the n-gram overlap with known routes.  
We use the bigram overlap ratio, i.e., we set $n$ to 2. 
Retro-BLEU penalizes routes with lengths exceeding $L$.
Given that the average length of patent routes is 2.79, we set $L$ to 3. 
In practice, we might have an estimation of the number of required reactions for each target molecule to set this value.

We compare Retro-BLEU with four other baselines: 
\begin{itemize}
    \item The route score by Badowski \textit{et al.}~\cite{costeffective2019}: This score takes into account route length and convergence. However, due to insufficient experimental data, the cost of each reaction and the yields can only be set using heuristics.
    We adapted the original implementation from PaRoutes in our comparisons\cite{genheden2022paroutes}:
    \begin{equation}
			Score_{\rm Badowski}(r) = \min_{\rm x\in pred(r)}{cost(x)}
       \end{equation}
    where $pred(r)$ denotes all the child nodes for the route $r$, \textit{i.e.,} the preceding reactions. The cost of a reaction $x$ is defined as:
    \begin{equation}
			cost(x) = \varepsilon(x) + \sum_{\rm r\in pred(x)}{\frac{cost(x)}{yield(x)}}
       \end{equation}
    where $\varepsilon(x)$ is the fixed reaction cost of performing the reaction. 
    Since both the reaction cost $\varepsilon(x)$ and the reaction yield $yield(x)$ are unknown, the authors heuristically set their values to 1 and 80\%, respectively.
    \item Cumulative probability: We recursively add the negative logarithmic probability obtained from the single-step retrosynthesis model NeuralSym\cite{NeuralSym2017} for each reaction in the route. 
    Note that for reactions in patent routes that cannot be predicted by the single-step model, we set its probability to 1e-10 when calculating the cumulative probability.
        \begin{equation}
			Score_{\rm cum}(r) = \sum_{x} \log{p(x)}
       \end{equation}
       where $x$ denotes each reaction in route $r$.
    \item Length: We use the number of reactions in the route as a metric, with shorter routes being preferable.
    \begin{equation}
			Score_{\rm length}(r) = N_{\rm x}(r)
       \end{equation}
    where $N_{\rm x}$ denotes the number of reactions in route $r$.
    \item Bigram Ratio: We rank the routes based on the bigram overlap ratio. As we discussed earlier, a higher bigram ratio suggests that the route more closely resembles known successful routes, and is therefore considered better.
    \end{itemize}

For each set of routes, we compute the route score using the aforementioned baselines and Retro-BLEU score. Then, we calculate the rank of the patent-recorded route among all the tested routes for the same target, leading to our top-k metric.
Since multiple routes may share the same scores (e.g., the same length under the route length metric), we assess the routes in terms of both best-case and worst-case scenarios. These scenarios represent instances where the patent route is identified either first or last among routes with the same score, respectively.

\section{Results and Discussions}

\subsection{Differentiating patent-extracted routes}

With Retro-BLEU, we can differentiate between plausible and implausible routes.
We report the results on the PaRoutes dataset.
We first merge patent routes with model-generated routes and assess their rankings using Retro-BLEU and other metrics in terms of top-k accuracy.
This evaluation method takes into account the ranking position of patent routes among the top-k.
We believe that validated routes derived from patents are chemically feasible and, as such, should receive higher scores (rank higher).
Although it is possible that model-generated routes can also be effective, experimentally evaluating numerous routes is impractical. 
Thus, we evaluate the ranking of patent routes as their feasibility is experimentally verified\cite{genheden2022paroutes}.
In addition, we refined a subset of set-n5 such that every reaction within each patent route in these subsets could be predicted among the top-50 predictions, we named these routes ``searchable routes'', resulting in 6,345 routes.
The reaction space in these routes more closely resembles that of the generated routes, and we consider this as a fairer comparison for utilizing the cumulative probability baseline.

\begin{figure*}[htbp]
\begin{center}        
\includegraphics[width=0.9\textwidth]{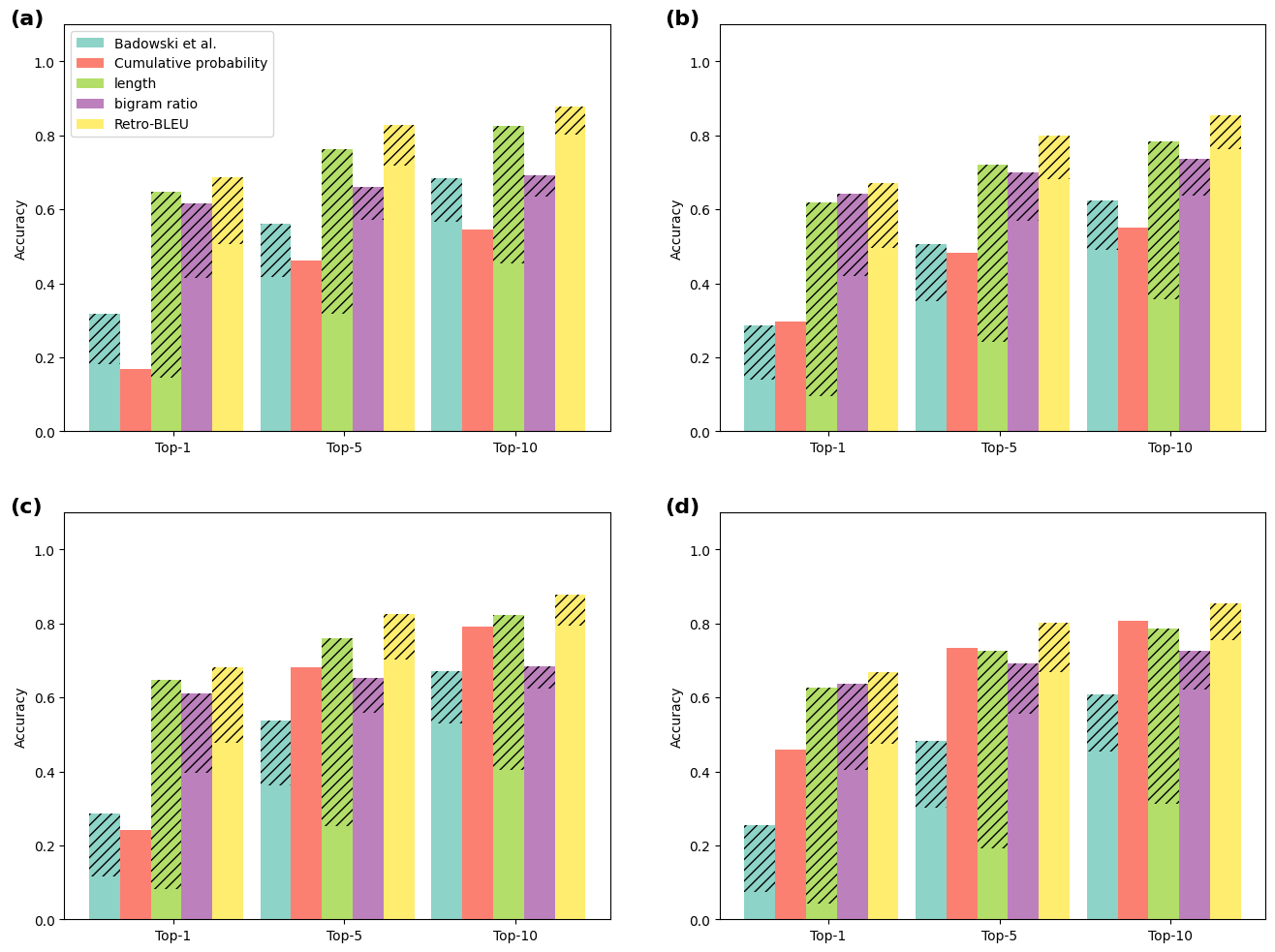}
\caption{Top-k accuracies for Retro-BLEU and other metrics. The top and bottom of areas with the diagonal line markings represent the best-case and worst-case scenarios, respectively. (a) MCTS algorithm applied to set-n5, (b) Retro$^*$ algorithm applied to set-n5, (c) MCTS algorithm applied to set-n5 searchable routes. (d) Retro$^*$ algorithm applied to set-n5 searchable routes.}
\label{fig2}
\end{center}
\end{figure*} 

Figure~\ref{fig2} shows the results on set-n5 for MCTS and Retro$^*$, and the results on set-n1 can be found in Supporting Figure 1, demonstrating a similar outcome to the one discussed here.
In Figure~\ref{fig2}, the gap between the best- and worst-case scenarios is marked with diagonal lines.
Retro-BLEU achieves the best overall ranking accuracy with a relatively small gap between the best- and worst-case when compared with other evaluation metrics on both MCTS and Retro$^*$ generated routes.

\begin{figure}[htbp]
\begin{center}
\includegraphics[width=0.4\textwidth]{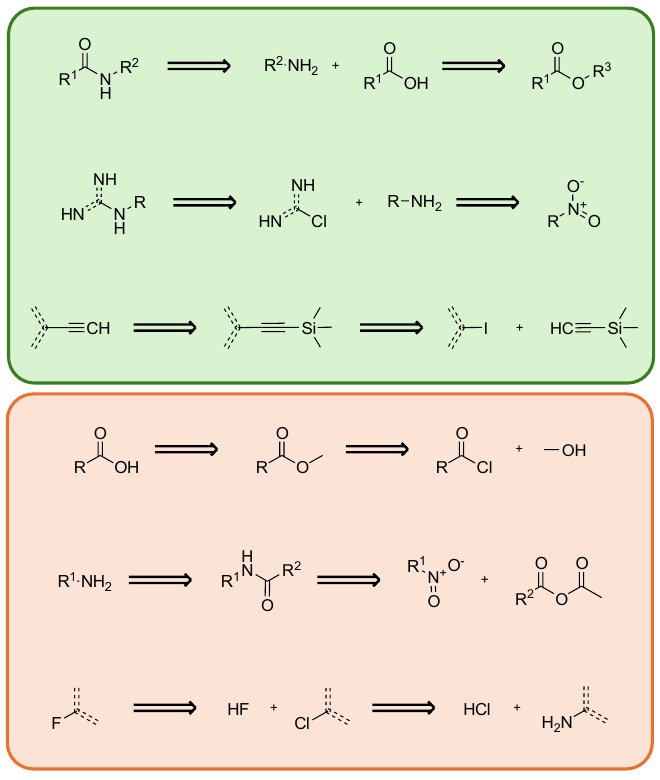}
\caption{Most frequent positive (highlighted in green) and negative (highlighted in red) template bigrams.}
\label{fig3}
\end{center}
\end{figure}

\subsection{Bigram examples}
Analyzing the extracted bigrams can help reveal the underlying correlations between bigrams and validated routes.
We analyzed some common template bigrams from patent routes, referred to as positive bigrams, and some common template bigrams extracted from generated routes that were not present in patent routes, referred to as negative bigrams.
We present three positive bigrams and three negative bigrams from the most frequently occurring bigrams  when planning routes for set-n5 molecules using MCTS in Figure~\ref{fig3}. 
Please note that the template bigrams presented here follow the retrosynthesis order, and we visualized them from the original auto-extracted SMARTS strings. 
This means that the template of the product is positioned to the left of the reaction sequence, and it is iteratively decomposed into the templates of the reactants on the right.

The first positive bigram illustrates a common process for generating an amide, which involves initially hydrolyzing an ester and then coupling it with another amine. 
Considering the difficulty of amidation reactions and the low reactivity of esters, hydrolyzing esters into more reactive carboxylic acids is often necessary.
Following this step, amidation can be completed with the help of condensing agents.
Similarly, in the second positive bigram, the product is deconstructed into a primary amine and a nitrogen-substituted heterocycle. 
The primary amine is derived from a nitro group through a reduction process.
This bigram demonstrates an excellent strategy for linking two molecular fragments together, which is commonly employed in drug-like molecule synthesis.
The third positive bigram comprises a Sonogashira coupling reaction followed by deprotection to form an exocyclic triple bond. 
Trimethylsilyl-based protection prevents the formation of side products from excessive coupling. 
The subsequent deprotection process provides an opportunity for coupling on the other side of the triple bond.
These positive bigrams represent well-established reaction strategies, whereas negative bigrams often contain redundant reactions that are not practical in synthesis applications.

In the first negative bigram, the overall reaction involves the hydrolysis of acyl chloride. 
However, the negative template bigram uses two steps to complete the entire process: an alcoholysis and a hydrolysis on the ester intermediate.
In common practice, this reaction can be simply executed by adding acyl chloride into water.
When performing the initial alcoholysis step, the search algorithm is unaware that the molecule will ultimately be converted into a carboxylic acid, leading to a redundant step.
The second negative bigram aims to convert the R-substituted nitro compound into a primary amine, which could be achieved by directly using reductants to reduce the nitro group. 
However, the template bigram incorporates extra reagents, resulting in an unnecessarily extended reaction sequence.
Similarly, the third negative bigram, which involves converting fluorobenzene to the more easily accessible aminobenzene, can be accomplished in a single step using the Schiemann reaction.
These negative bigrams reveal an inherent limitation in the current retrosynthesis planning approaches, in which consecutive reactions were not considered, consequently resulting in the potential generation of redundant steps. 
We can potentially build these negative bigrams using various data mining techniques, which can help in early stopping unnecessary searches during route finding.

\begin{figure*}[htbp]
\begin{center}
\includegraphics[width=\textwidth]{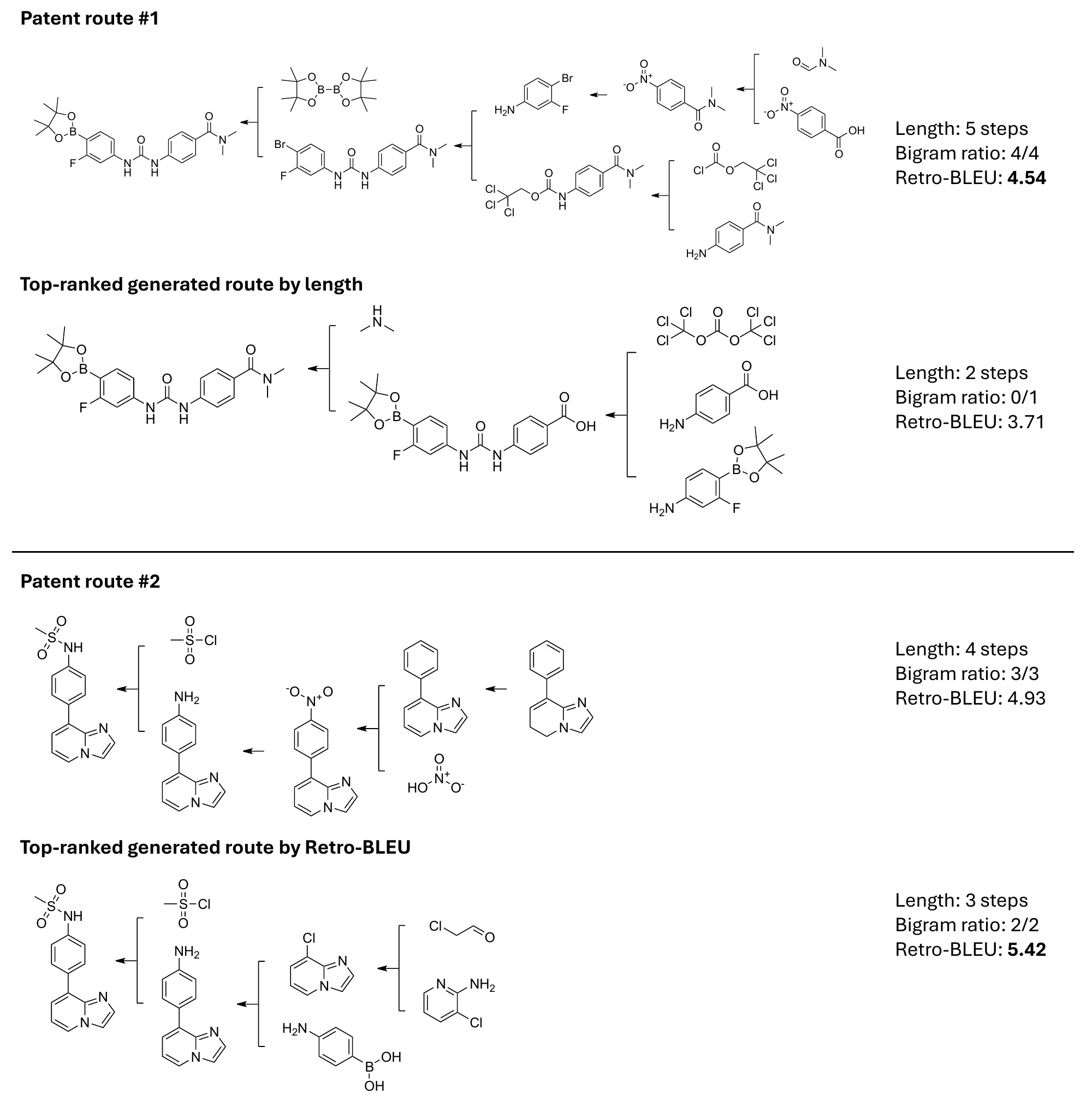}
\caption{Examples of using Retro-BLEU to select feasible retrosynthesis routes. Top: The recorded route in patent WO2012/58671 and the shortest generated route. Bottom: The recorded route in patent US4596872 and the top-ranked generated route ranked by Retro-BLEU. The search algorithm employed in these examples is MCTS.}
\label{fig4}
\end{center}
\end{figure*} 

\subsection{Synthesis route examples}
We show two cases to further verify the correlation between Retro-BLEU and chemical plausibility in Figure ~\ref{fig4}.
Figure~\ref{fig4}(a) compares a relatively long route (5 steps) from the patent database with the shortest one generated by AiZynthFinder using the MCTS algorithm. 
The patent route synthesized the target molecule in 5 steps, starting from simple molecules, and featured a convergent route. 
It incorporated several synthesis strategies 
such as linking two acyl amines together step-by-step using a 2,2,2-trichloroethyl chloroformate. 
All four template bigrams have been previously documented, resulting in a Retro-BLEU score of 4.54.
The generated route shown in Figure~\ref{fig4}(a) illustrates an unfeasible approach due to potential selectivity issues in the second step. 
This route attempts to combine two components using bis(trichloromethyl) carbonate in a single step. 
The presence of two identical reactive trichloromethyl groups in bis(trichloromethyl) carbonate may cause the molecule to couple twice, resulting in overreacted by-products.
The template for this three-component reaction is quite rare and is likely only applicable to specific reactants or accompanied by subsequent deprotection reactions. 
However, in this generated route, the next step involves the coupling of carboxylic acid and secondary amine, a reaction sequence that has not been recorded.
In this case, the shortest path score is only 3.71, which is significantly lower than the corresponding patent route.
While many searching algorithms aim to finding shorter routes, the length can be deceptive and leading to unhelpful routes for synthesis scientists.
Although this generated route consists of only two steps, it cannot synthesize the expected molecule in an actual laboratory environment.
The difference in Retro-BLEU scores demonstrates that our metric is capable of identifying potential invalid template combination patterns. 
This enables preliminary screening of synthetic routes and assists scientists in avoiding unfeasible approaches.

It is worth mentioning that for approximately 20\% of target molecules, the patent-extracted routes are not ranked in the top-1 positions.
We selected another example where the Retro-BLEU score of the patent route is lower than some of the generated routes, as shown in~\ref{fig4}(b).
The patent route synthesizes the target molecule within four steps, primarily modifying the substituent on the benzene ring; however, the starting material remains relatively expensive\footnote{The price was \$280/g on https://www.biosynth.com/ accessed in July, 2023.}.  
In our comparison, we selected the generated route with the highest Retro-BLEU score (5.42), which surpasses the patent route's score of 4.84. 
Notably, this route was originally ranked in the 25$^{\rm th}$ position by AiZynthFinder's searching process. 
The template bigrams in the generated route have all been recorded, and it can be considered an alternative synthesis route by first synthesizing the two aromatic systems and then coupling them together using a Suzuki-Miyaura coupling reaction. 
The generated route is shorter than the one previously reported in the patent, and the starting materials are also simplified.
This comparison indicates that model-generated routes can also serve as valuable supplements to existing routes if carefully selected based on Retro-BLEU scores.
Therefore, we believe that Retro-BLEU can serve as a valuable metric to distinguish plausible routes from a vast number of model-generated routes, ultimately enhancing the efficiency and effectiveness of synthetic route selection.

\section{Conclusion}
Retro-BLEU, as a statistical metric, has its limitations. 
Although our study confirms a correlation between Retro-BLEU and chemical feasibility, it should be noted that correlation does not imply causation. 
Because Retro-BLEU does not explicitly encode chemical knowledge, it might face difficulties in evaluating routes involving rare or novel reactions due to their limited occurrences in the database, potentially underestimating innovative routes.
Additional factors, such as production rate and reaction costs, among others, could also be considered when evaluating a synthesis route, given that sufficient data is available.
The limitation can be mitigated in the future by incorporating more data in constructing the template sequence database to ensure a comprehensive and diverse representation of reactions.
Furthermore, it is crucial to regularly update the template sequence database with the latest research findings and innovative reactions. 
This ensures that the metric remains relevant and effective in evaluating state-of-the-art retrosynthetic routes, thereby maintaining its ability to identify and assess novel synthesis pathways in the rapidly evolving field of chemistry.

In conclusion, we introduce Retro-BLEU as a metric to evaluate and rank retrosynthetic routes generated by computer-aided synthesis planning tools.
Retro-BLEU offers a statistical approach to assess chemical plausibility by analyzing reaction template sequences, and it significantly outperforms other baselines in selecting experimentally validated patent routes. 
This accelerates the utilization of retrosynthesis planning tools and enables researchers to identify feasible routes more efficiently. 
We encourage further research for evaluating model-generated retrosynthetic routes, which will support synthetic chemistry progress and facilitate the discovery and synthesis of novel molecules, benefiting the broader scientific community.

\section{Data and Software Availability}

We primarily utilized data from PaRoutes to conduct our research. 
PaRoutes is an open-source benchmark that can be accessed at \url{https://github.com/MolecularAI/PaRoutes}. 
Additionally, we employed AiZynthFinder and Retro* for our retrosynthesis planning, which are other common open-source tools in this field. 
They can be accessed at \url{https://github.com/MolecularAI/aizynthfinder} and \url{https://github.com/binghong-ml/retro_star}, respectively. 

The code for Retro-BLEU is publicly available at \url{https://github.com/catalystforyou/Retro-BLEU}. 

\begin{acknowledgement}

The authors thank Dr. Kangjie Lin for helpful discussions.

\end{acknowledgement}

\begin{suppinfo}

The following files are available free of charge.
\begin{itemize}
  \item Reaction/Template n-gram overlap analysis under different partition settings, template n-gram overlap analysis under different radii, and the relationship between Retro-BLEU and filtering strategies.
\end{itemize}

\end{suppinfo}

\bibliography{achemso-demo}

\end{document}